# Optimizing Neuro-Fuzzy and Colonial Competition Algorithms for Skin Cancer Diagnosis in Dermatoscopic Images

1st Hamideh Khaleghpour
*Department of Computer Science*
*The University of Tulsa*
Tulsa, United States
hamideh-khaleghpour@utulsa.edu*

2nd Brett. McKinney
*Department of Computer Science*
*The University of Tulsa*
Tulsa, United States
brett-mckinney@utulsa.edu

***Abstract*— The rising incidence of skin cancer, coupled with limited public awareness and a shortfall in clinical expertise, underscores an urgent need for advanced diagnostic aids. Artificial Intelligence (AI) has emerged as a promising tool in this domain, particularly for distinguishing malignant from benign skin lesions. Leveraging publicly available datasets of skin lesions, researchers have been developing AI-based diagnostic solutions. However, the integration of such computer systems in clinical settings is still nascent. This study aims to bridge this gap by employing a fusion of image processing techniques and machine learning algorithms, specifically neuro-fuzzy and colonial competition approaches. Applied to dermoscopic images from the ISIC database, our method achieved a notable accuracy of 94% on a dataset of 560 images. These results underscore the potential of our approach in aiding clinicians in the early detection of melanoma, thereby contributing significantly to skin cancer diagnostics.**

## I. Introduction

The skin, the human body's largest organ, serves critical functions including protection against heat, sunlight, wounds, and infections. It regulates body temperature, stores water and fat, and synthesizes vitamin D. Structurally, the skin comprises three main layers: the epidermis (surface layer), dermis (inner layer), and hypodermis [1].

Skin cancer, particularly melanoma, ranks among the deadliest cancers, with its incidence rising globally. Early detection significantly increases the chances of recovery, emphasizing the need for methods that facilitate prompt diagnosis. Melanoma, a lethal melanocytic tumor arising from abnormal melanin production, accounts for approximately 50% of skin cancer-related deaths. Originating in the epidermal and dermal layers, melanoma manifests through the proliferation of melanin spots, spreading to the skin's surface. Clinically, the spread of these lesions and their differentiation from the background are crucial diagnostic indicators [2].

The criticality of early skin cancer detection extends beyond melanoma, as it increases the risk of other cancers. However, early-stage melanoma diagnosis remains challenging, even for specialists. In response, interdisciplinary efforts in dermatology and artificial intelligence aim to develop computer systems capable of early melanoma screening [3].

This study develops an intelligent system aimed at early detection of melanoma through the analysis of dermatoscopic images. The system is engineered to discern and analyze key characteristics such as color, diameter, density, and shape. It utilizes sophisticated algorithms, including the colonial competition algorithm, renowned for its effectiveness in precise lesion categorization. A central feature of this research is the use of a median filter, which significantly improves the distinction between lesions and normal skin. This enhancement is critical in aiding medical professionals to accurately identify melanoma in its initial stages, thereby potentially improving patient survival rates.

Recent AI-based diagnostic and anomaly detection studies emphasize that high model performance alone is not sufficient for real-world deployment. Robust preprocessing, interpretable feature extraction, leakage-safe evaluation, and reliable probability estimation are essential when machine learning models are used in high-stakes domains such as healthcare and cybersecurity [20], [21].

The objective of this research is to engineer a system capable of identifying early-stage melanoma lesions. This is achieved by capturing and analyzing images of the affected skin, extracting crucial features such as color, size, and structural properties. The analysis focuses on reducing these characteristics to a manageable dataset, thereby facilitating swift and accurate classification of the lesions. The adoption of various algorithms, particularly the colonial competition algorithm, is a testament to the system's innovative approach. The method employed is non-invasive, relying predominantly on visual assessment, which, despite its challenges, offers a significant advantage in early melanoma detection. Precise differentiation between the lesion and its background is crucial for accurate diagnosis. The median filter plays a pivotal role in this process, enhancing the clarity of the lesion's boundaries. Subsequently, an advanced demarcation algorithm is applied to the optimally chosen image, ensuring the most accurate distinction between the melanoma lesion and the surrounding skin [4].

## II. Theoretical Foundations

### A. Skin Cancer

Skin cancer, a prevalent form of cancer globally, accounts for about 75% of all cancer diagnoses. It is marked by abnormal growths in the skin's outermost layer. Despite its high cure rate, skin cancer remains a significant health concern due to its widespread occurrence. Preventive strategies include limiting exposure to ultraviolet rays, a known risk factor. Skin cancer can manifest on any part of the body and is not limited to any specific skin color. Common types include basal cell carcinoma, squamous cell carcinoma, and melanoma [5].

### B. Colonial Competition Algorithm

In the realm of optimization algorithms, nature-inspired approaches have gained prominence. Examples include genetic algorithms, ant colony optimization, and simulated annealing. The Colonial Competition Algorithm, a novel addition, diverges from natural inspirations, instead drawing from the socio-political process of human colonization. This algorithm models the dynamics of colonization for optimization purposes, proving effective in both static and dynamic contexts, as well as constrained and unconstrained optimization scenarios. Originally designed for continuous problems, its adaptations now successfully address discrete optimization challenges [6].

### C. Neuro-Fuzzy System

The Adaptive Neuro-Fuzzy Inference System (ANFIS) merges neural networks with fuzzy logic, enabling it to harness the strengths of both in a unified framework. Primarily governed by fuzzy if-then rules, ANFIS excels in approximating non-linear functions, making it a robust estimator. Its structure resembles Radial Basis Function neural networks, and gradient descent algorithms are employed for parameter optimization, a process known as neuro-fuzzy modeling. ANFIS leverages both empirical data and foundational knowledge for system creation [7].

### D. K-Means Segmentation Algorithm

Introduced by McQueen, the K-means algorithm is an unsupervised classification technique that organizes data into predefined classes, minimizing an error function. It is widely recognized for its efficacy in feature space partitioning [8].

### E. Ant Colony Algorithm

The Ant Colony Optimization algorithm, inspired by the foraging behavior of ant colonies, optimizes routes from nests to food sources. This meta-heuristic method uses pheromone trails as a model for algorithmic optimization, capturing the collective intelligence of ants in finding optimal solutions.

## III. Advancements in Skin Cancer Diagnosis: A Review of Computational Methods

In recent years, significant strides have been made in the field of skin cancer diagnosis using computational methods. In 2023, Naghavipour, AI-Nahari and Zandi did a comparative study was conducted between three methods which are Artificial Neural Network (ANN), Support Vector Machine (SVM) and K-Nearest Neighbor (KNN) to help to diagnose melanoma by extracting features from dermatoscopic images and their classification. Their results show that ANN could achieve better accuracy (83.5%) [9].

In 2019, Fujisawa et al. leveraged deep learning techniques, a subset of machine learning, for this purpose [2]. The following year, Hussain et al. explored skin cancer management and prevention strategies. Their work, pivotal in the context of skin cancer's rising prevalence and severe implications, has become a valuable resource for researchers focusing on image processing and skin cancer diagnosis, particularly in the use of image processing techniques for melanoma detection [10].

In 2017, Jana et al. introduced novel computer analyses and filters for melanoma diagnosis, providing a comprehensive overview of the methodologies prevalent in contemporary research [11]. A year earlier, in 2016, Joseph et al. developed innovative algorithms based on morphology operations, which involve processing images using geometric shapes. This approach involves comparing the similarity of shapes in an image with a geometric constructive element, utilizing a dataset of 97 skin lesion images and implementing the algorithm through Python Programming [12].

Lau et al.'s 2009 study presented a method to classify skin lesions as benign or malignant, employing artificial neural networks. Utilizing standard cameras to capture skin lesion images, their initial experiments demonstrated a promising classification accuracy of 90.71%, markedly outperforming previously documented methods [13].

In the same vein, 2019 saw Manickavasagam and Selvan propose an automated lung cancer classification method using CT scan images, combining neuro-fuzzy logic with the cuckoo algorithm [14]. Concurrently, Maron et al. applied a convolutional neural network algorithm to diagnose skin cancer [15].

Mhaske and Phalke in 2013 utilized the SVM (Support Vector Machine) method for melanoma image classification. Their technique involved converting images from RGB to grayscale and applying a classifier; pixels exceeding a certain threshold were identified as lesions, while lower values indicated normal skin [16].

Rezvantalab et al., in 2018, employed neural networks and deep learning for melanoma diagnosis and identification [17], and Rashad and Takruri in 2016 introduced an automated melanoma diagnosis method based on the support vector algorithm [18].

Khaleghpour and McKinney proposed a leakage-safe graph-feature framework for fraud detection in temporal transaction networks, highlighting the importance of interpretable feature engineering, strict evaluation protocols, and operationally meaningful metrics beyond accuracy. Although their work focuses on financial fraud rather than medical imaging, its methodological emphasis on interpretability and reliable evaluation is relevant to AI-based diagnostic systems. In another study, Khaleghpour and McKinney introduced a unified AI framework for audio anomaly detection, combining preprocessing, feature extraction, and machine learning models to improve robustness in noisy and complex data environments [20], [21].

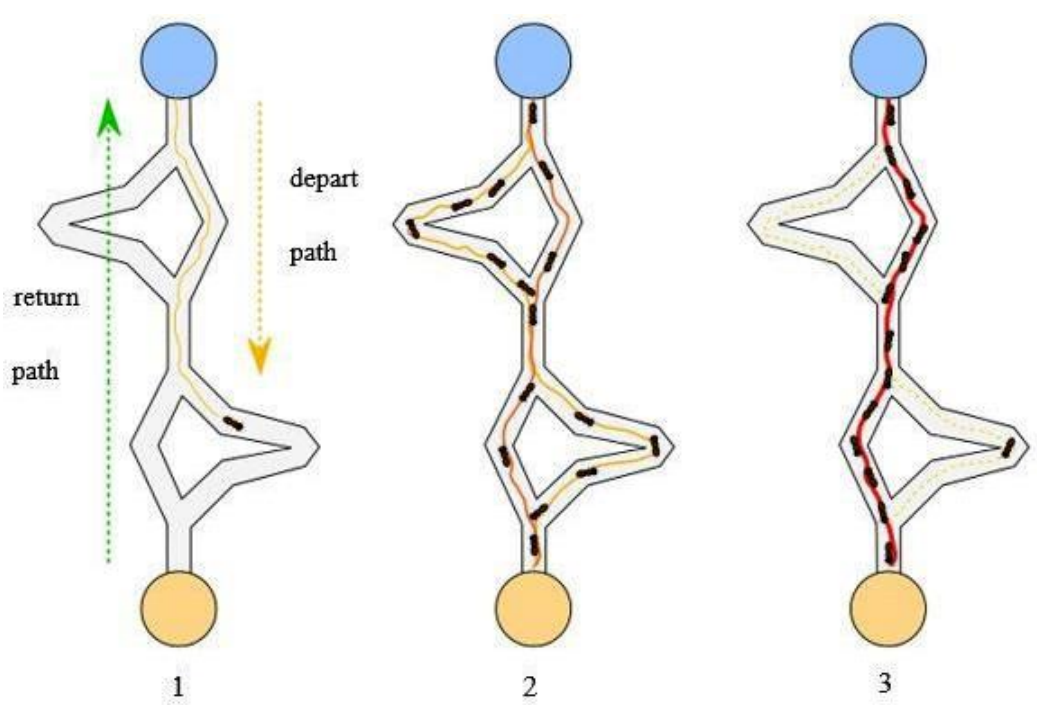

Fig. 1. The ant's round trip route [10]

Finally, in 2016, Satheesha et al. explored novel methods to reduce feature dimensions in their research. Their unique approach, applied to a dataset of 320 dermatoscope images, involved using a median filter to reduce noise and employing SVM for classification. The dataset included 100 healthy and 220 diseased images, with the latter comprising 100 melanocytic moles and 120 melanomas [19]. Dermatoscopes, crucial in dermatology for enabling up to 10x magnification and natural light, aid in the accurate and timely diagnosis of various skin conditions.

## IV. Research Methodology

### A. Proposed Algorithm Overview

This section presents an overview of the proposed algorithm, detailing its fundamental components and the sequential stages involved, as visualized in the subsequent figure. The algorithm's process initiates with the input of captured images and progresses through several crucial phases before culminating in segmentation and classification. The procedure encompasses the following steps:

- Image Registration: This preliminary phase involves the standardization and alignment of the input images, setting the stage for uniform processing across the entire dataset.
- Image Preprocessing: At this juncture, the images undergo a series of preparatory processing techniques. These steps are designed to enhance image quality and are crucial for subsequent analysis, including but not limited to noise reduction, contrast adjustment, and color standardization.
- Image Segmentation: The segmentation phase divides the images into distinct regions or segments. This crucial step isolates significant features from the non-essential background, facilitating focused analysis.
- Feature Extraction: This stage is dedicated to identifying and extracting key characteristics and attributes from the segmented images, laying the groundwork for the next phase of classification.
- Classification with Neuro-Fuzzy-Colonial Competition: The final step employs a sophisticated neuro-fuzzy-colonial competition approach for classification. This method combines the adaptive strengths of neuro-fuzzy systems with the strategic optimization of colonial competition algorithms, aiming for high-precision categorization.

Each stage of the algorithm is meticulously crafted to ensure both the integrity and effectiveness of the analysis, thus guaranteeing precise and dependable results in the segmentation and classification of images.

## V. Data Analysis

### A. Database Utilization

Our research leveraged the widely recognized ISIC global standard database, which is also commonly used by other researchers in this field. The ISIC Melanoma Project primarily aims to enhance melanoma diagnosis' accuracy and efficiency, thereby reducing mortality rates and unnecessary biopsies. This initiative includes developing digital imaging standards and compiling a public archive of clinical and dermoscopic skin lesion images.

### B. Method of Implementation

Early detection of melanoma, a deadly skin cancer type, is crucial. Medical professionals often use the ABCD method for early melanoma identification, where each letter represents a different characteristic of a mole or skin lesion. In our research, these characteristics extracted from images are utilized in the neuro-fuzzy training phase using the colonial competition algorithm. It is important to note that while most skin moles, brown spots, and lesions are harmless, they can occasionally indicate melanoma. Individuals with more than 100 moles are at higher risk. Early melanoma signs often manifest in one or two unusual moles, highlighting the importance of regular skin checks and recognizing changes in moles. Key features analyzed include:

- A (Asymmetry): Benign moles are typically symmetrical. A line drawn through the middle of a melanoma mole reveals asymmetry, indicating a potential concern.
- B (Border): Non-cancerous moles have smooth, even borders, whereas melanoma moles may exhibit uneven, notched, or scalloped edges.
- C (Color): Uniform brown coloration characterizes benign moles. Variation in color, including different shades of brown, black, red, white, or blue, can signify melanoma.
- D (Diameter and Size): Commonly, benign moles are smaller than malignant ones. Melanoma lesions are often larger than a pencil eraser (about 6 mm) but can be smaller when first detected.

The implementation follows the flowchart outlined in the previous chapter:

- First Step (Image Registration): We processed 560 images from the ISIC database.

- Second Step (Pre-processing): Images were standardized to 500x500 pixels to expedite processing. A median filter was applied for contrast enhancement and noise reduction.
- Third Step (Image Segmentation): After median filtering, we employed K-means segmentation algorithms and thresholding to isolate lesions from healthy areas. Subsequently, image features were extracted based on the ABCD rules.

Below, Fig.2 showcases a selection of the original images utilized in our research.

- Fourth Step (Neuro-Fuzzy - Colonial Competition Implementation): We analyzed 13 characteristics, including lesion diameter, sphericity, color spectrum, and edge uniformity. The data was divided into training and testing sets for implementation and evaluation of the neuro-fuzzy machine learning algorithm trained by the colonial competition algorithm.

### C. Algorithm Evaluation

In dataset categorization, achieving high accuracy and precision is paramount, particularly in medical diagnoses. For instance, while false negatives (healthy individuals diagnosed as sick) might be tolerable to some extent, false positives (sick individuals missed) are unacceptable. The Confusion Matrix assists in evaluating such scenarios by categorizing each data sample into one of four states: True Positive (TP), False Negative (FN), True Negative (TN), and False Positive (FP). The performance of each category can be assessed using the following formulae:

$$Accuracy = (TP + TN) / (TP + FN + FP + TN) \quad (1)$$

While accuracy is a fundamental metric, it's not the sole criterion for evaluation. Sensitivity (or True Positive Rate), which indicates the proportion of correctly identified positive cases, is also crucial:

$$Sensitivity\ (TPR) = TP / (TP + FN) \quad (2)$$

### D. Research Data Insights

We compared our proposed algorithm against two neuro-fuzzy algorithms and a combined neuro-fuzzy and ant colony algorithm. The sensitivity and accuracy of these algorithms were analyzed for effectiveness. Unlike reference studies using around 100 images, our research utilized 560 images, testing the algorithm's robustness. The data were split into training and testing groups in a 70:30 ratio. Optimal parameter values, determined through trial and error and aligned with those used in other studies, were employed in our Python-based implementations.

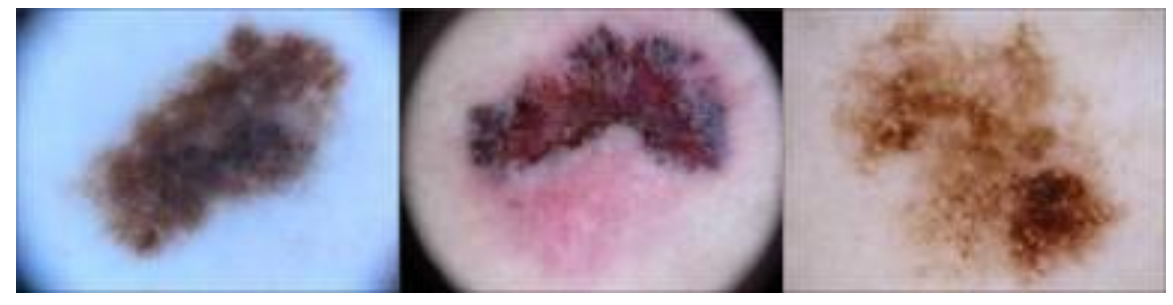

Fig. 2. Selection of Original Images

TABLE I. TABLE OF IMPLEMENTATION PARAMETERS

| Algorithm | parameters |
|---|---|
| The proposed algorithm | Number of repetitions: 200<br>Number of empires: 5<br>Population of colonial competition: 200<br>Coefficient of revolution: 0.1 |
| neuro-fuzzy algorithm | Number of repetitions: 200 |
| Algorithm of neuro-fuzzy and ant colony | Number of repetitions: 200<br>The number of ant population: 20 |

TABLE II. A SAMPLE OF FEATURES

| | | | | | | | | | |
|---|---|---|---|---|---|---|---|---|---|
| 1 | 64/15905 | 0/574186 | 0/993547 | 0/88246 | 0/8926 | 11/36519 | 14/45734 | 21/0563 | 2/755188 |
| 2 | 285/9167 | 0/837802 | 0/968912 | 0/531741 | 0/926439 | 50/07578 | 45/08184 | 53/43263 | 29/45372 |
| 1 | 6/579525 | 0/632036 | 0/978571 | 0/496843 | 0/95518 | 32/13093 | 44/29128 | 62/52592 | 19/51421 |
| 2 | 192/0301 | 0/87926 | 0/9641 | 0/343483 | 0/991254 | 90/05452 | 78/42911 | 98/82362 | 75/43867 |
| 1 | 259/5444 | 0/877663 | 0/954588 | 0/212466 | 0/986195 | 75/75457 | 71/30104 | 72/60035 | 117/4308 |
| 1 | 285/5491 | 0/83865 | 0/953476 | 0/299341 | 0/958479 | 54/36637 | 44/57605 | 38/0532 | 162/2957 |

The table II presents a sample of features extracted during the image processing phase, specifically from the initial 13 images obtained.

The extracted features for each image in the dataset encompass various aspects, including the irregularity index, asymmetry degree, variance in red, green, and blue color channels, ratios of red, green, and blue, as well as differences in brightness, colorlessness, and lesion color. In the dataset, the last column categorizes the images, with '1' indicating a normal mole and '2' signifying melanoma. Notably, color characteristics, particularly the red color variance, are among the most reliable indicators for skin cancer diagnosis. Thus, these features are crucial for accurate skin cancer detection. The subsequent table illustrates the effectiveness of different classification methods applied to the aforementioned database.

The results presented below clearly indicate that the proposed combined algorithm demonstrates superior performance on both the test and training datasets.

TABLE III. SENSITIVITY RESULTS

| Method | Training data sensitivity results | test data sensitivity results |
| --- | --- | --- |
| The proposed algorithm | %93 | %90 |
| neuro-fuzzy algorithm | %90 | %81 |
| neuro-fuzzy algorithm and ant colony | %83 | %81 |

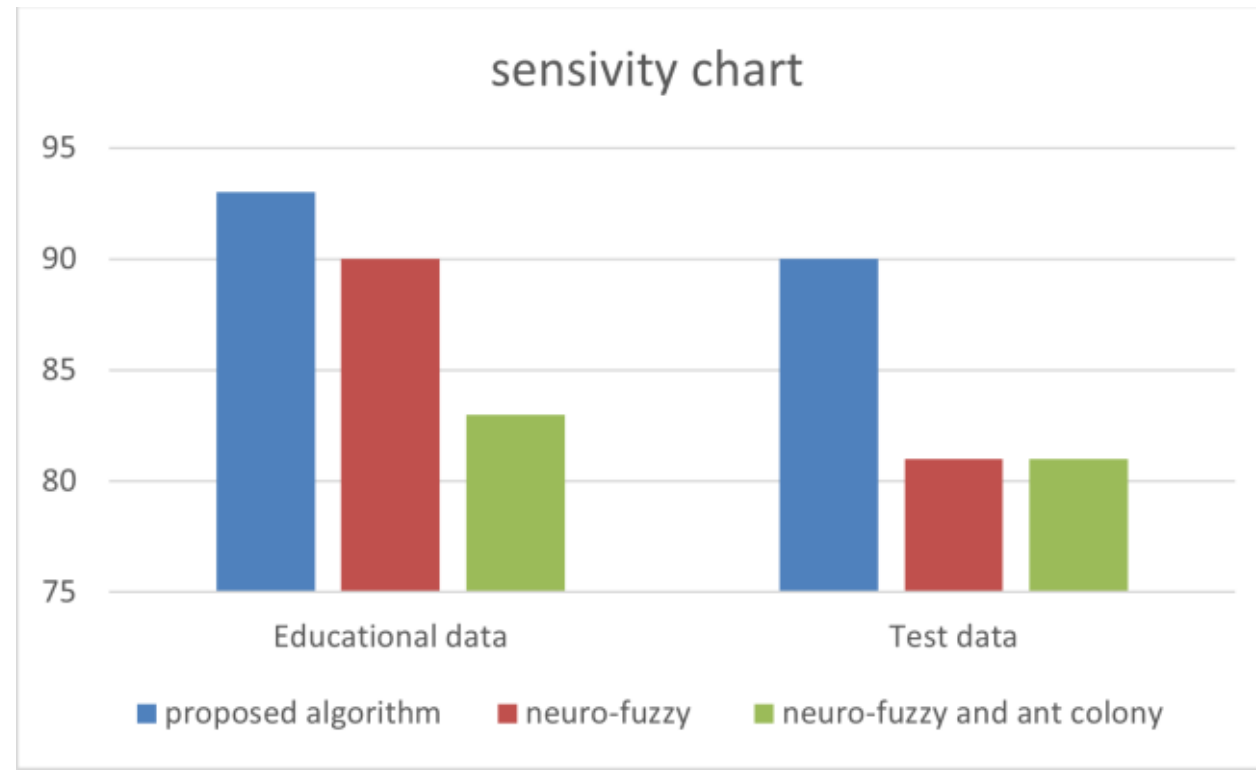

Fig. 3. Results of the sensitivity graph

TABLE IV. ACCURACY RESULTS

| Method | Validation results of educational data | Test data accuracy results |
| --- | --- | --- |
| The proposed algorithm | %97 | %94 |
| neuro-fuzzy algorithm | %95 | %86 |
| neuro-fuzzy algorithm and ant colony | %88 | %86 |

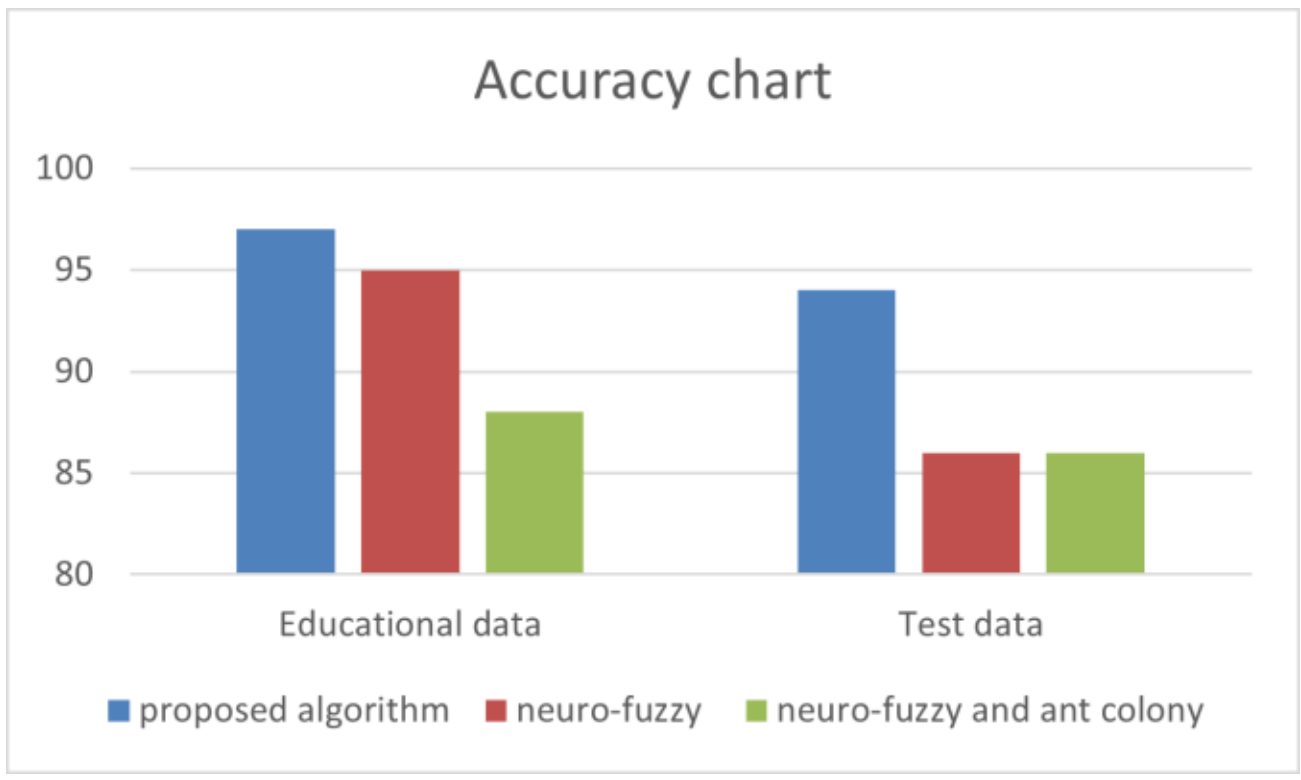

Fig. 4. Accuracy graph

The integration of neuro-fuzzy and colonial competition techniques offers several benefits, notably rapid convergence and the capability to optimize functions with numerous variables. Evident from the outcomes, this approach yields more favorable results in both testing and training datasets compared to the other two methods examined.

### *E. Convergence Results of the Proposed Algorithm*

To evaluate the algorithm's convergence, the proposed method was applied to varying sets of images – 50, 100, 200, 300, and 400 – each undergoing different numbers of repetitions: 50, 100, and 200. This process was designed to thoroughly assess the algorithm's performance across diverse conditions.

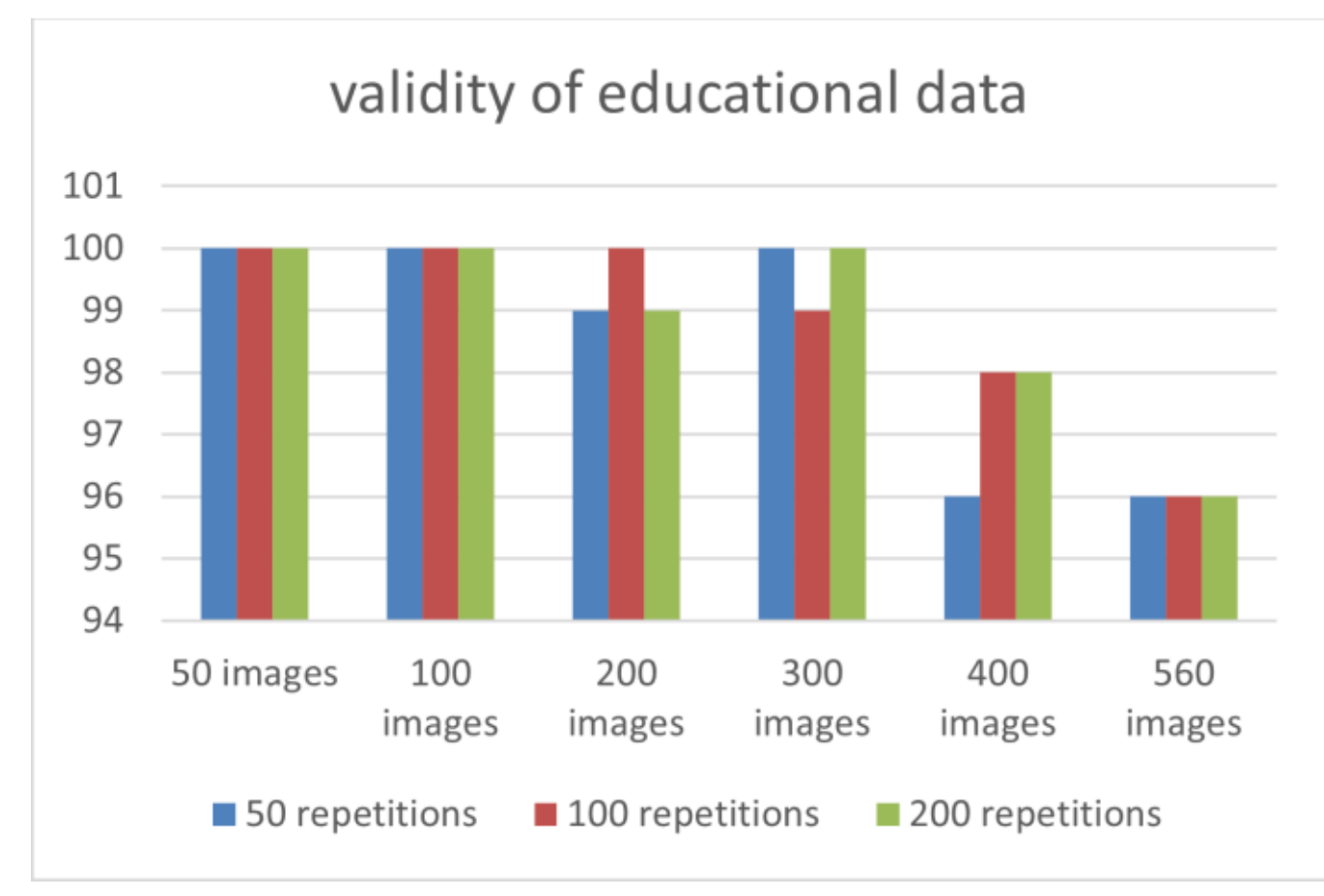

Fig. 5. Convergence accuracy results of the training data

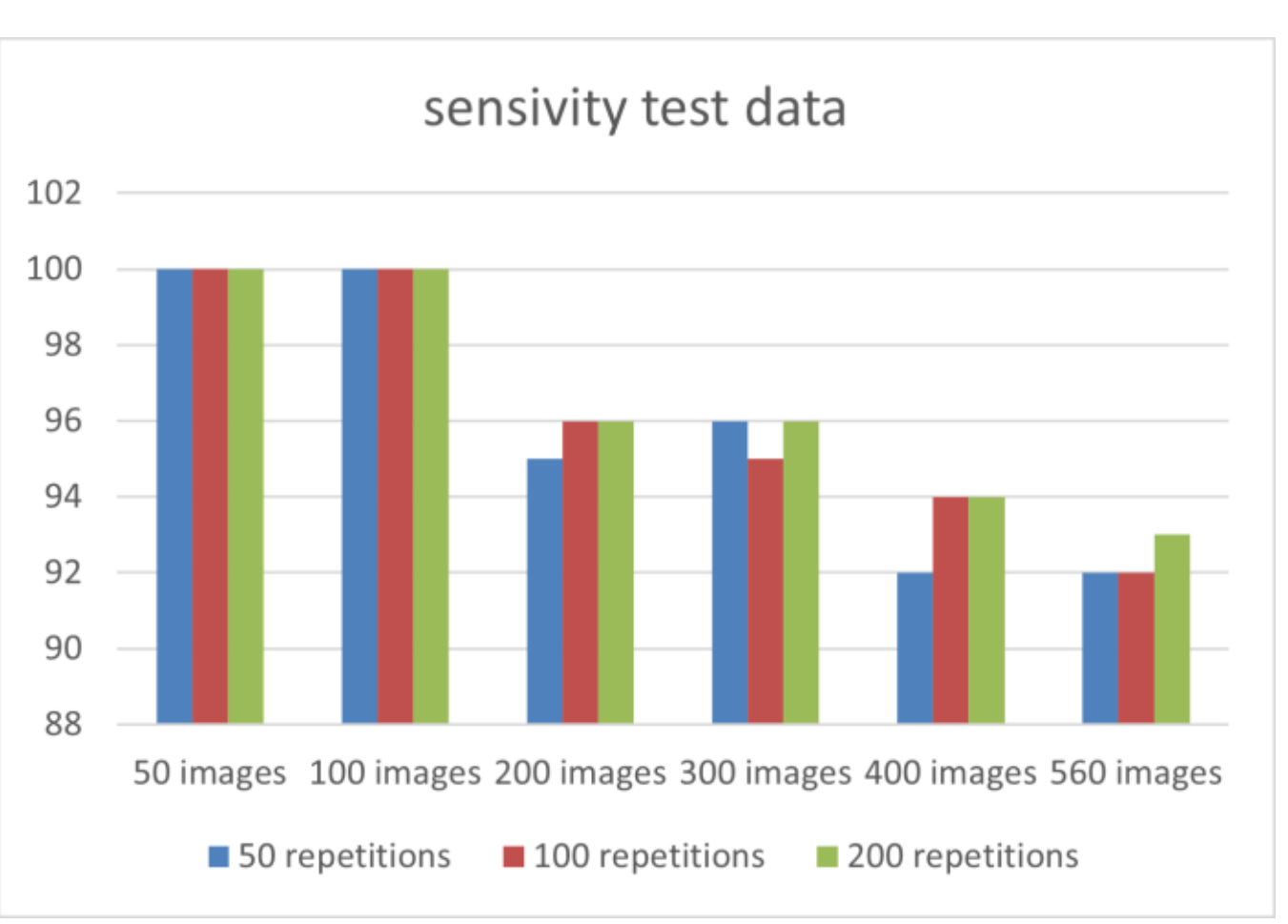

Fig. 6. The results of the sensitivity of training data convergence

It is observable that with an increase in the number of images, the proposed algorithm continues to deliver satisfactory outcomes. Furthermore, with a greater number of iterations, the performance improves, attributable to enhanced training.

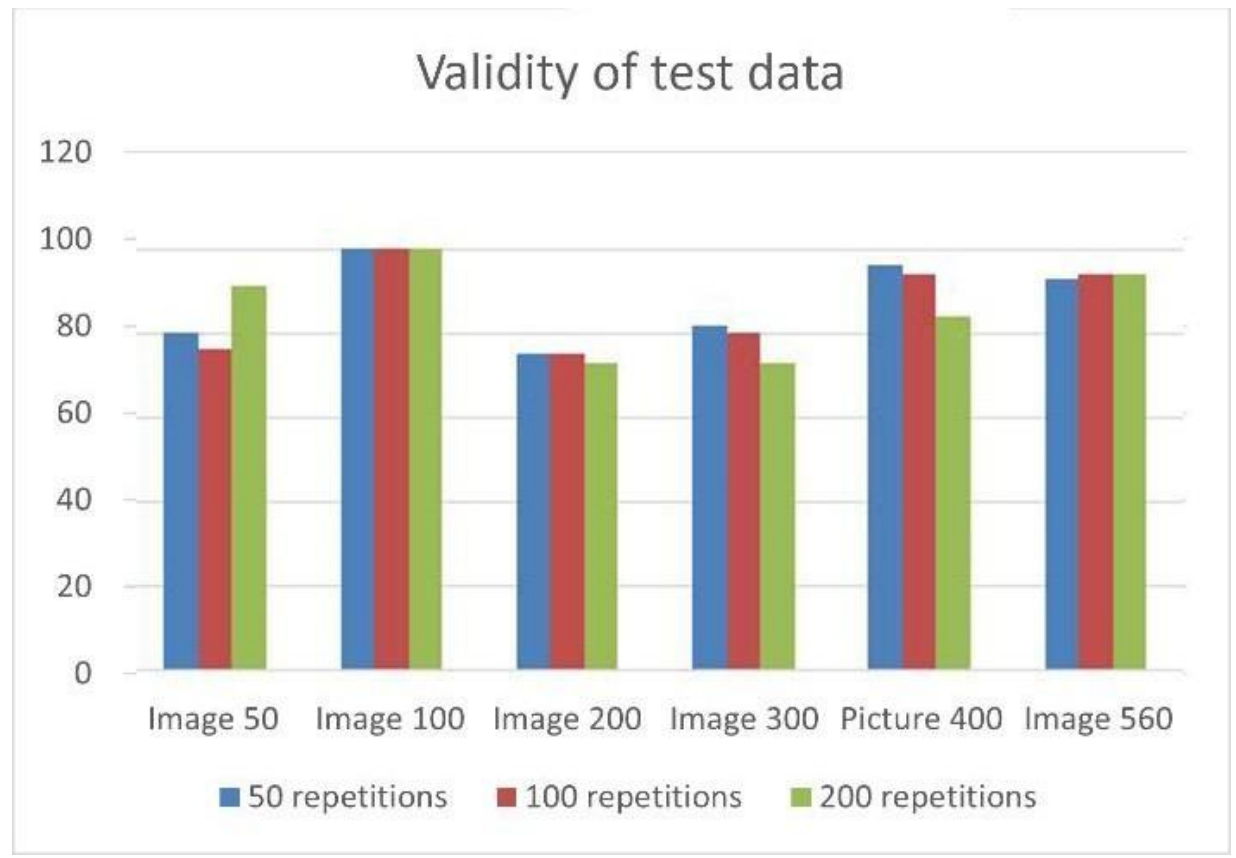

Fig. 7. Test data convergence accuracy results

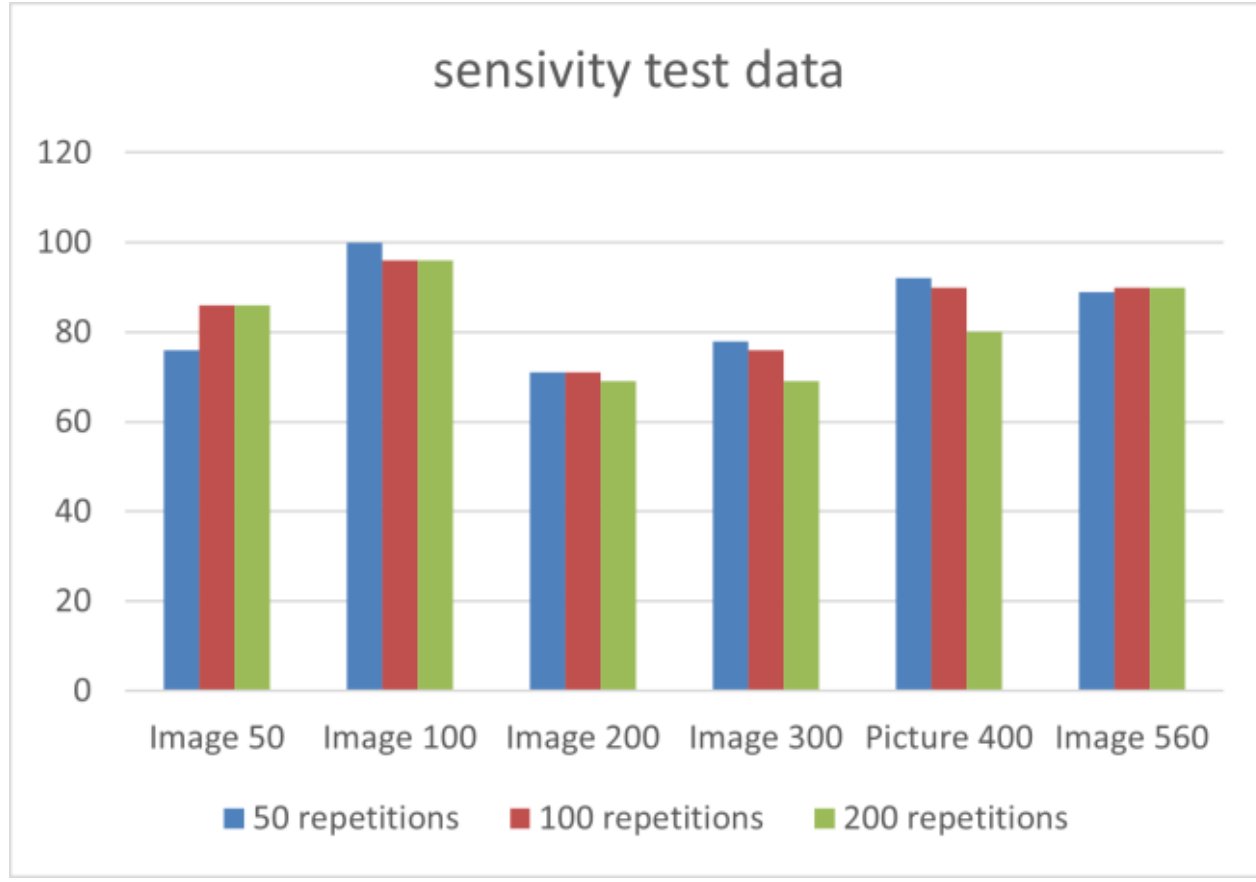

Fig. 8. Test data sensitivity convergence results

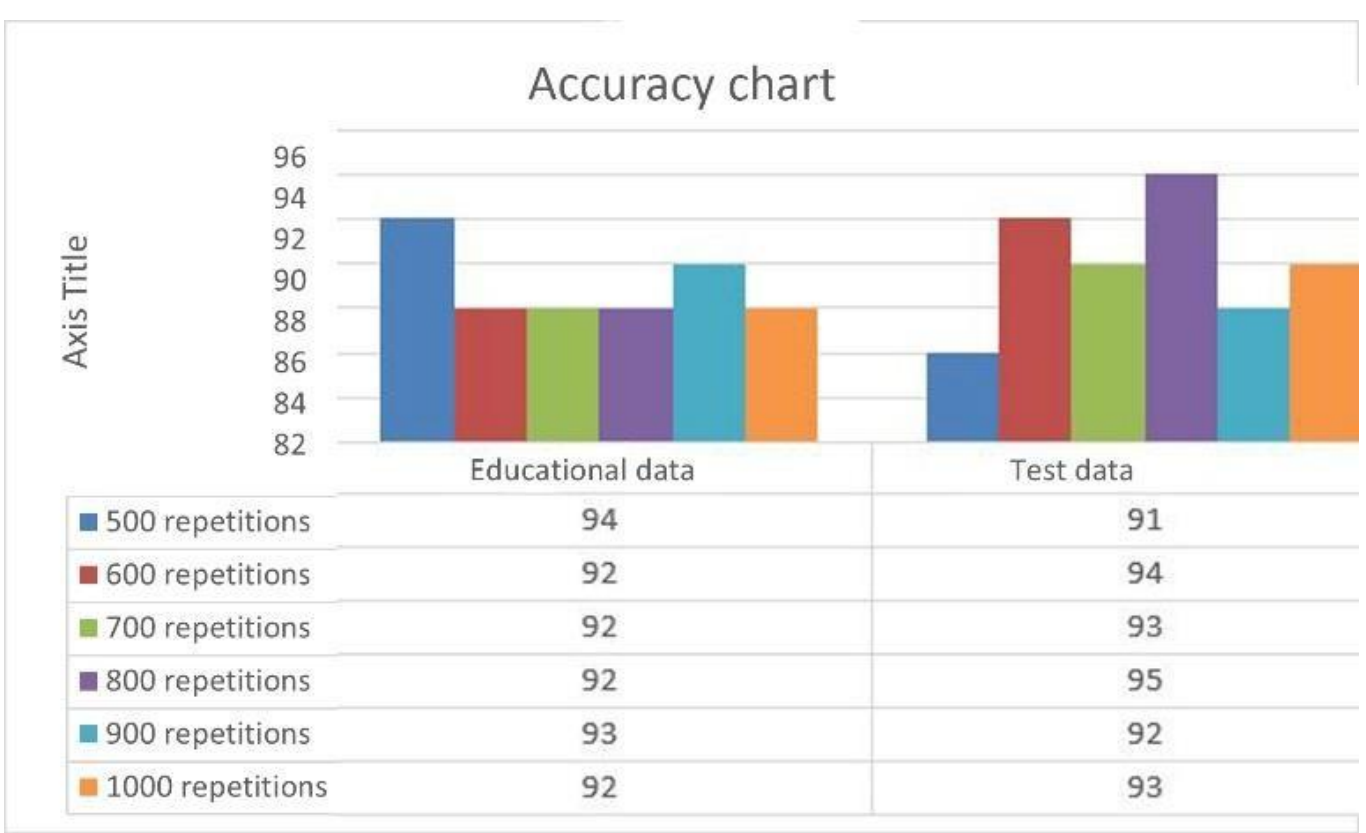

Fig. 9. Accuracy graph with different iterations

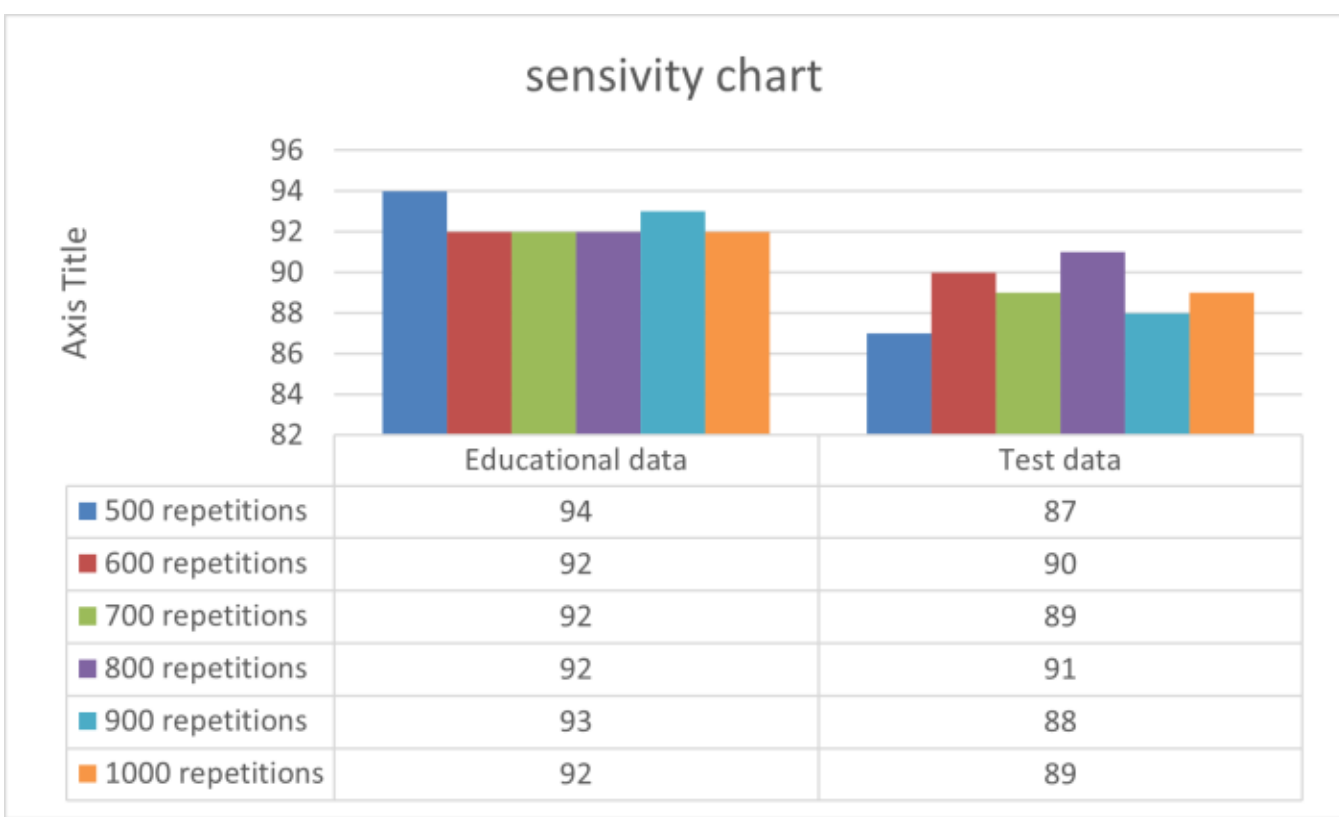

Fig. 10. Sensitivity graph with different iterations

As the results showed, the number of higher images is challenging and the number of repetitions is effective in the results, from 500 repetitions to 1000 repetitions on 560 images.

## VI. Conclusion

This study introduces an innovative method for the intelligent diagnosis of melanoma skin cancer, utilizing a synergistic approach combining image processing and a neuro-fuzzy-colonial competition algorithm. The effectiveness of this method, particularly in terms of accuracy and sensitivity, surpasses several existing methodologies in diagnosing this type of lesion. With millions diagnosed with skin cancer annually, and the necessity of extensive testing for diagnosis, this method stands out for its rapid detection capabilities, potentially enhancing survival rates.

The development of the proposed system was methodically segmented into distinct phases. During the pre-processing stage, we explored and implemented techniques to enhance image quality. A median filter was employed to not only improve the quality but also to bring clarity to image details. Subsequently, segmentation was executed to minimize computational time and focus processing efforts. Here, the K-means method and thresholding were utilized, drawing from extensive research in medical image segmentation. The final step involved feeding extracted image features, such as lesion sphericity, color, and area, into a machine learning algorithm for classification. This process was then benchmarked against neuro-fuzzy methods and a neuro-fuzzy-ant colony algorithm combination. Our system demonstrated high accuracy and efficiency compared to existing models.

In our research, we particularly investigated the impact of the median filter in conjunction with the neuro-fuzzy-colonial competition method for melanoma identification, yielding promising results. The implementation trials were conducted on the ISIC database, comprising 560 images, under consistent conditions across all three algorithms. As detailed in Chapter 4, the ant colony algorithm, while adjustable, tends to converge quickly to an optimal solution, resulting in slightly inferior outcomes compared to the colonial competition algorithm. Our proposed algorithm exhibited superior performance on training data, outperforming the other methods in terms of convergence and accuracy on test data, irrespective of the number of repetitions or varying image counts.